\documentclass{article}
\usepackage{amssymb}
\usepackage{IEEEtrantools}
\usepackage{booktabs} 
\usepackage{multirow}
\usepackage{fontawesome}
\usepackage{tikz}
\usepackage{makecell}   
\usepackage{spconf,amsmath,graphicx,hyperref}

\title{Adaptive Frequency Domain Alignment Network \\ for Medical image segmentation}
\name{Zhanwei Li, Liang Li\sthanks{Corresponding author: \href{mailto:liangli@tju.edu.cn}{\faEnvelope~ liangli@tju.edu.cn}}  and Jiawan Zhang}
\address{
    College of Intelligence and Computing, Tianjin University \\
    Tianjin 300350, PR China \\
}

\begin{document}
%
\maketitle
\begin{abstract}
High-quality annotated data plays a crucial role in achieving accurate segmentation. However, such data for medical image segmentation are often scarce due to the time-consuming and labor-intensive nature of manual annotation. To address this challenge, we propose the \emph{Adaptive Frequency Domain Alignment Network (AFDAN)}—a novel domain adaptation framework designed to align features in the frequency domain and alleviate data scarcity. AFDAN integrates three core components to enable robust cross-domain knowledge transfer: an \emph{Adversarial Domain Learning Module} that transfers features from the source to the target domain; a \emph{Source-Target Frequency Fusion Module} that blends frequency representations across domains; and a \emph{Spatial-Frequency Integration Module} that combines both frequency and spatial features to further enhance segmentation accuracy across domains. Extensive experiments demonstrate the effectiveness of AFDAN: it achieves an Intersection over Union (IoU) of 90.9\% for vitiligo segmentation in the newly constructed VITILIGO2025 dataset and a competitive IoU of 82.6\% on the retinal vessel segmentation benchmark DRIVE, surpassing existing state-of-the-art approaches. 

\end{abstract}
\begin{keywords}domain adaptation, frequency alignment, adversarial learning, medical image segmentation
\end{keywords}
\section{Introduction}

For large-scale medical image segmentation, high-quality annotated data are essential for achieving accurate model performance. However, due to the heavy reliance on expert domain knowledge and the substantial costs of annotation, such data remain scarce in most practical scenarios. This scarcity severely constrains model generalization, particularly in target domains characterized by complex data distributions or limited sample availability \cite{11094586}.

To address these limitations, transferring knowledge from existing annotated datasets in related domains has emerged as a valuable research direction\cite{DANN2016}. However, most current approaches operate under the assumption that the source and target domains share identical or highly similar annotation categories. For instance, Jeon \emph{et al.} \cite{SciRep2024} directly apply models pre-trained on public datasets to multi-organ segmentation tasks. In real-world clinical settings, such assumptions rarely hold. To illustrate, the source domain may contain only annotations of normal anatomical structures, whereas the target task requires the segmentation of pathological regions, or the source may include multi-class annotations while the target focuses on only a subset. The \emph{annotation discrepancy} significantly complicates knowledge transfer across domains, frequently leading to negative transfer or the failure of domain adaptation.

\def\colSep{1.15cm}
\def\rowSep{1.15cm}
\begin{figure}
    \centering
    \begin{tikzpicture}
        \node[inner sep=0] (image1) at (0,0) {\includegraphics[width=.24\linewidth]{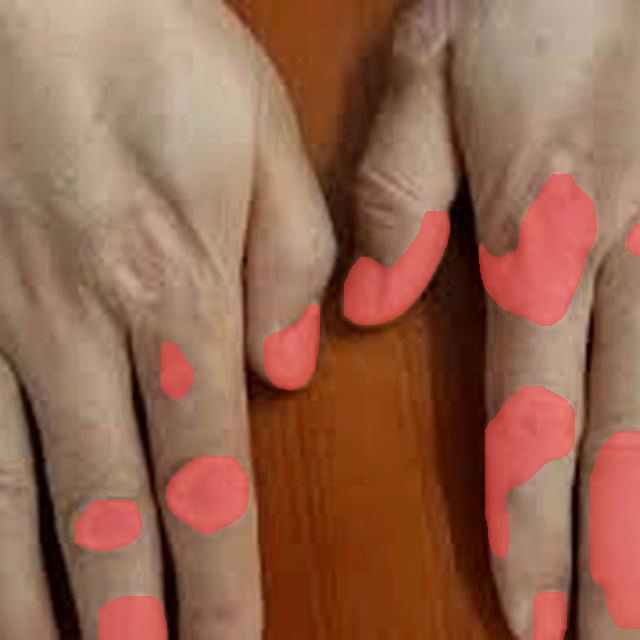}};
        \node[inner sep=0, right of=image1, xshift=\colSep] (image2) {\includegraphics[width=.24\linewidth]{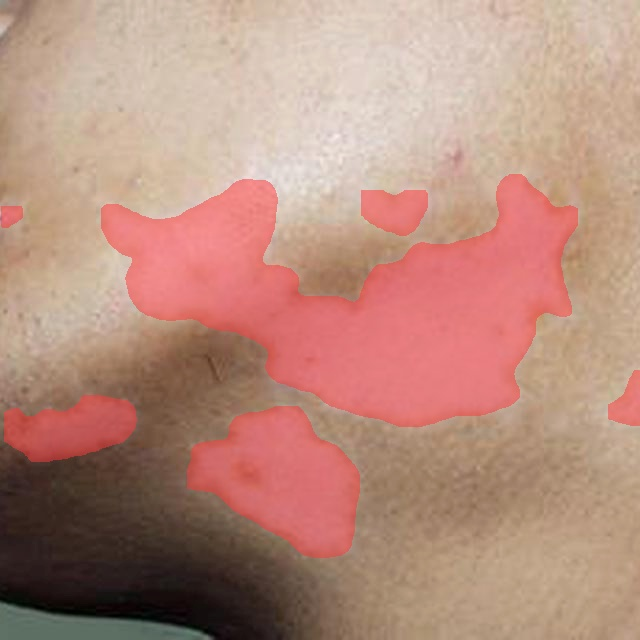}};
        \node[inner sep=0, right of=image2, xshift=\colSep] (image3) {\includegraphics[width=.24\linewidth]{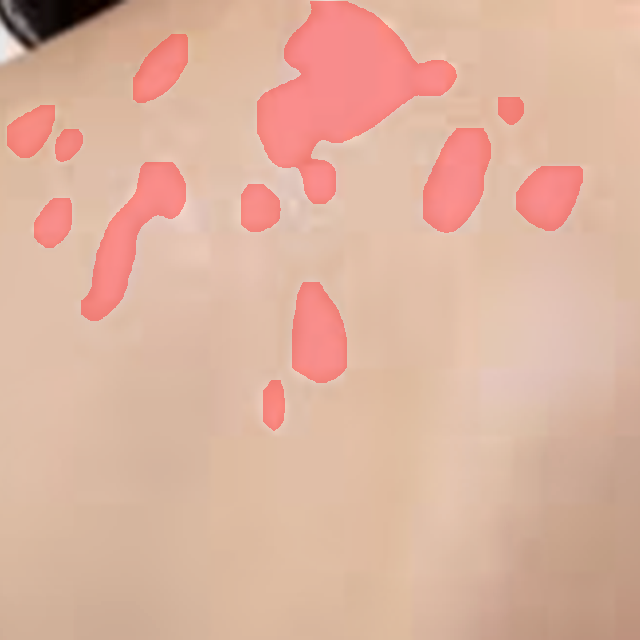}};
        \node[inner sep=0, right of=image3, xshift=\colSep] (image4) {\includegraphics[width=.24\linewidth]{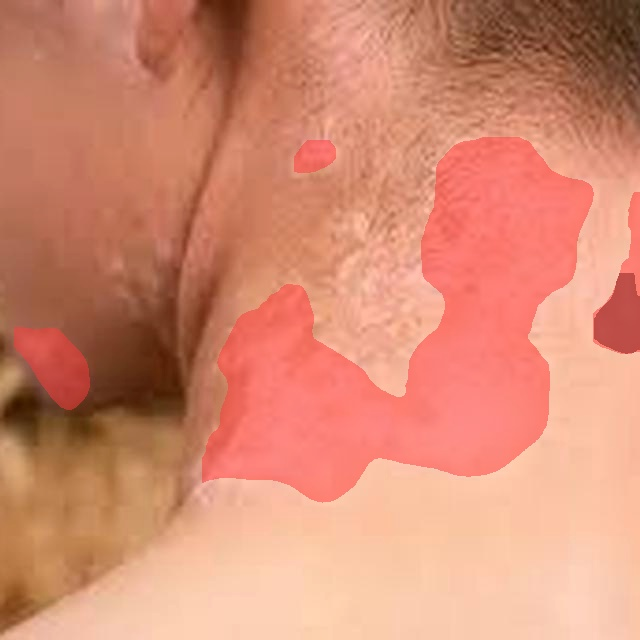}};
        
        \node[inner sep=0, below of=image1, yshift=-\rowSep] (image5) {\includegraphics[width=.24\linewidth]{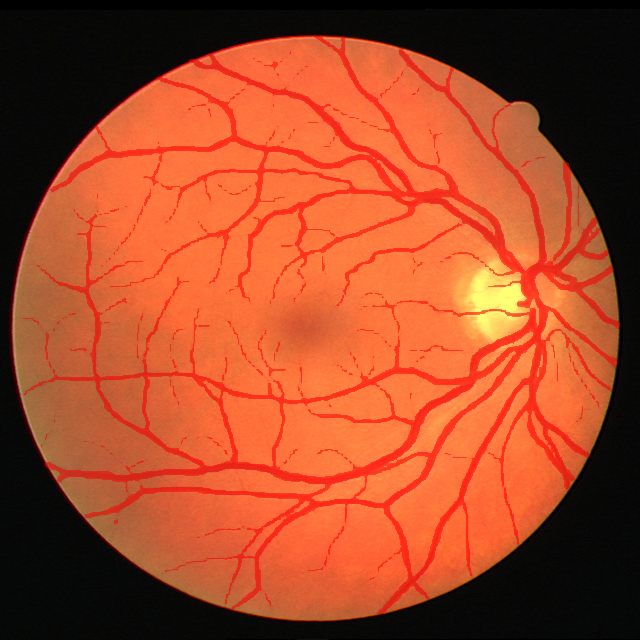}};
        \node[inner sep=0, below of=image2, yshift=-\rowSep] (image6) {\includegraphics[width=.24\linewidth]{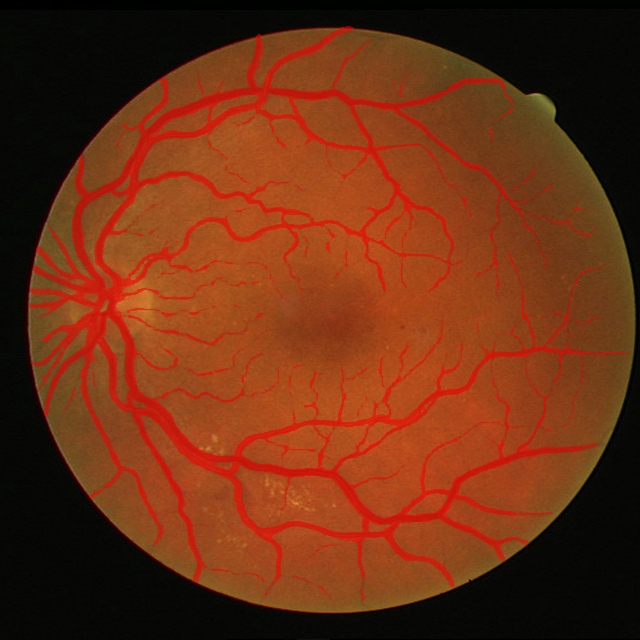}};
        \node[inner sep=0, below of=image3, yshift=-\rowSep] (image7) {\includegraphics[width=.24\linewidth]{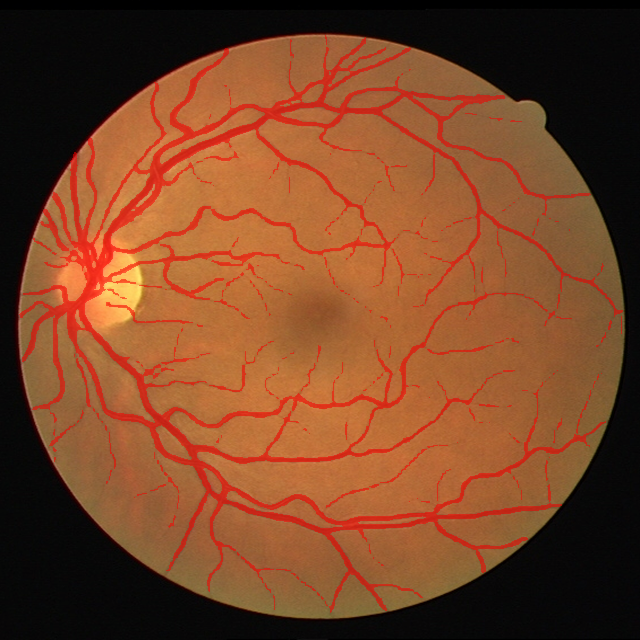}};
        \node[inner sep=0, below of=image4, yshift=-\rowSep] (image8) {\includegraphics[width=.24\linewidth]{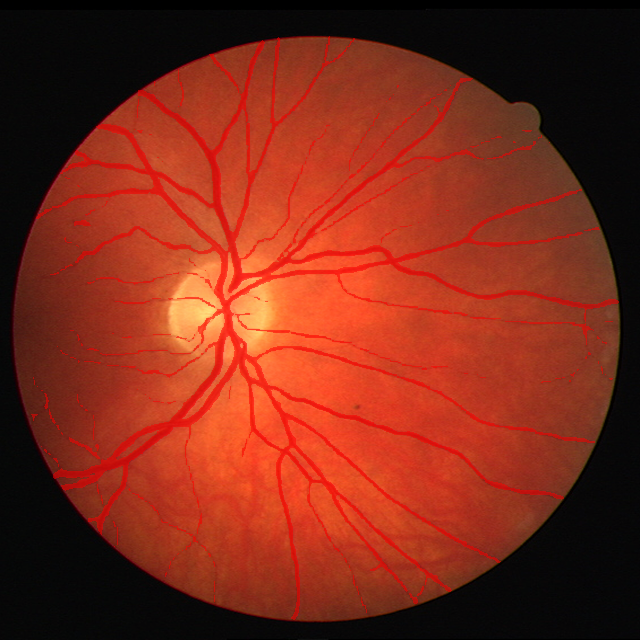}};
        \node[] at (-0.05,-0.8) {\color{white} Vitiligo Seg.};
        \node[] at (-0.05,-3.0) {\color{white} Vessels  Seg.};
    \end{tikzpicture}
    \caption{Segmentation results for vitiligo (first row) and retinal vessels (second row). Predicted regions are highlighted in red (best viewed in color). Note that both results were obtained without direct training on datasets annotated specifically for these target structures: the vitiligo segmentation model was trained using skin lesion annotations, while the retinal vessel segmentation was derived from a model trained with general fundus image annotations.}
    \label{fig:placeholder}
\end{figure}
Recent studies \cite{FDA2020,trans2023} have sought to mitigate inter-domain distribution gaps through methods such as partial feature alignment\cite{fu2021partial}, pseudo-label generation\cite{yang2024consistency}, and label mapping\cite{Chen2023Frequency}. Nonetheless, most of these approaches do not adequately resolve the deeper semantic gap introduced by mismatched annotation spaces \cite{Dapan}. Developing transfer learning frameworks that account for annotation disparities to enable effective knowledge transfer and semantic structural alignment remains a critical yet underexplored challenge.

By decoupling structural (phase) and semantic (amplitude) information, the frequency-domain approach minimizes its impact on the semantic content of images\cite{huang2021fsdr}.   This inherent property makes it less susceptible to negative transfer and more focused on capturing domain-invariant representations.
In this work, we address the challenge of annotation discrepancies in medical image segmentation by introducing the \emph{Adaptive Frequency Domain Alignment Network (AFDAN)}. Our contributions are as follows.
\begin{enumerate}
    \item We propose AFDAN, a novel cross-domain segmentation network. It integrates three key modules: the Adversarial Domain Learning (ADL) module , the Source-Target Frequency Fusion (STFF) module, and the Spatial-Frequency Integration (SFI) module. Specifically, the ADL module generates target-specific frequency features from source domain images to enable effective cross-domain alignment; the STFF module accelerates the construction of target domain frequency features, thereby improving both efficiency and adaptability; and the SFI module combines complementary representations from the frequency and spatial domains to further enhance segmentation accuracy and preserve fine anatomical structures. Collectively, these components reduce domain shift, preserve structural detail, and improve the detection of subtle lesions.
    \item We conduct comprehensive experiments across multiple benchmarks. AFDAN achieves state-of-the-art performance, with an Intersection over Union (IoU) of 90.9\% on the newly constructed VITILIGO2025 dataset \cite{Guo2022} and 82.6\% on the DRIVE \cite{DRIVE} retinal vessel segmentation benchmark. As illustrated in Fig. \ref{fig:placeholder}, the segmentation outputs exhibit high visual fidelity, accurately delineating structural details across both normal and pathological regions.
\end{enumerate}

\section{Method}
\label{sec:pagestyle}

\begin{figure*}[ht]
\centering 
\includegraphics[width=\textwidth]{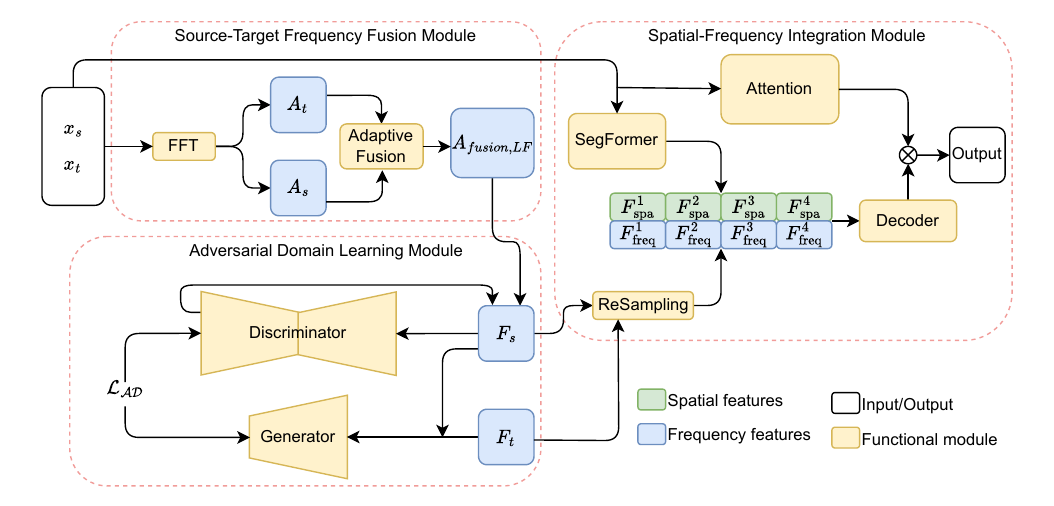}
\caption{The source and target domain images, $x_s$ and $x_t$, are first transformed into the frequency domain via Fast Fourier transform (FFT). An Adaptive Fusion mechanism is then applied to accelerate domain adaptation. The second module, the Adversarial Domain Learning module, reduces domain shift, where $F_t$ denotes the target-domain features. Finally, the spatial and frequency features are jointly fed into the Spatial-Frequency Integration module. The integrated features are then concatenated, passed through the decoder, and multiplied with the attention map to produce the segmentation result, enabling effective integration of spatial and frequency features.}
\label{model_image}
\end{figure*}

This section introduces three core modules of the proposed model AFDAN. The overall architecture of AFDAN and the relationship between three modules is shown in Fig. \ref{model_image}.

\subsection{Adversarial Domain Learning Module} 

Inspired by DANN \cite{DANN2016}, we employ an Adversarial Domain Learning module to align features from source and target domains, as shown in Fig. \ref{model_image}. Given the observation that amplitude spectrum contains domain-specific style characteristics \cite{pasta2023}, we propose a frequency-aware adversarial learning strategy specifically designed for amplitude spectrum alignment. 
Denoting the discriminator and generator as $D$ and $G$ respectively, the adversarial loss is defined as:
\begin{IEEEeqnarray}{c}
\mathcal{L}_{\text{AD}} = \mathbb{E}_{A_t} \left[ \log D(A_t) \right] + \mathbb{E}_{A_s} \left[ \log \left( 1 - D(G(A_s)) \right) \right],
\label{eq:adv_loss}
\end{IEEEeqnarray}
where $A_s$ and $A_t$ denote the amplitude spectra of the source and target domains, respectively. Optimizing this adversarial objective encourages the generator $G$ to produce amplitude representations $G(A_s)$ that are indistinguishable from $A_t$ to the discriminator $D$, thereby effectively aligning the source amplitude distribution with the target domain. Notably, our Adversarial Domain Learning module plays an indispensable role in alleviating annotation discrepancies between the source and target domains.


\subsection{Source-Target Frequency Fusion Module}

Motivated by the need to accelerate domain adaptation and prevent overfitting, we introduce a Source–Target Frequency Fusion module. Fourier analysis indicates that the phase spectrum primarily encodes structural information, while the amplitude spectrum governs intensity distribution—a finding leveraged in domain adaptation to improve structural consistency across domains \cite{leedomain}. Our module preserves the source phase to retain critical structures (e.g., lesion contours) and aligns only the amplitude across domains, thereby maintaining anatomical integrity while adapting appearance \cite{zhang2022cyclemix}.

For source-domain features $F_s \in \mathbb{R}^{B\times C\times H\times W}$ from $x_s$ and target-domain features $F_t \in \mathbb{R}^{B\times C\times H\times W}$ from $x_t$, where $x_s$ denotes source-domain images and $x_t$ denotes target-domain images, we decompose them via 2D FFT as:
\begin{equation}
F_s = A_s e^{i\Phi_s}, \quad F_t = A_t e^{i\Phi_t},
\end{equation}
where $A_s, A_t$ represent amplitude spectra from both domains and $\Phi_s, \Phi_t \in [-\pi,\pi]^{B\times C\times H\times W}$ denote phase spectra .

To align intensity distributions without compromising anatomical integrity, we directly retain the source domain’s phase spectrum to preserve structural consistency. For amplitude alignment while retaining their respective high-frequency components to maintain important structural details \cite{luo2024fourier}, we fuse the low-frequency components of the source and target amplitude spectra via:
\begin{IEEEeqnarray}{c}
A_{\text{fusion, LF}} = \alpha \cdot A_s + (1-\alpha) \cdot A_t,
\end{IEEEeqnarray}
where $\alpha \sim \mathcal{U}(0,1)$ is a random mixing coefficient drawn from a uniform distribution. This stochastic fusion strategy constructs dense intermediate amplitude distributions between domains, thereby effectively facilitating the reduction of annotation discrepancies.


\subsection{Spatial-Frequency Integration Module}

Our efficient architecture integrates spatial-frequency features, which is critical for capturing both global context and local edge details \cite{Texlivernet}. To handle spatial dimension mismatches between frequency features ($F_{\text{freq}}^k$, where $k=1,2,3,4$, corresponding to features with distinct output dimensions) and spatial features ($F_{\text{spa}}^k$), we use bilinear interpolation to resample $F_{\text{freq}}^k$ to match the size of $F_{\text{spa}}^k$ while preserving spatial relationships. The aligned features are then concatenated channel-wise into $F_{\text{cat}}^k$, effectively combining complementary frequency-based structural cues and spatial semantic information.

An attention mechanism is further introduced to suppress specialized noise (e.g., device artifacts) and highlight domain-invariant regions \cite{MeGACDA}. The concatenated features are processed by a decoder to recover structural and contextual details, after which the output is modulated via element-wise multiplication with a spatial attention map to produce the final segmentation. This design enhances discriminative power and emphasizes task-relevant patterns, improving overall robustness and accuracy. As illustrated in Fig. \ref{model_image}, our framework leverages the modular and replaceable design of SegFormer \cite{segformer}, allowing flexible component selection based on task complexity and dataset scale.

\section{Experiment}
\label{sec:typestyle}

\subsection{Dataset} 

To tackle annotation scarcity in medical image segmentation, we propose a cross-domain training approach evaluated through two complementary experiments.

For the vitiligo segmentation task, we first adopt a cross-domain setup: the ISIC2018 dataset \cite{isic2018} serves as the source domain, providing dermatoscopic images with high-quality masks across multiple skin diseases to enable robust feature learning. The target domain consists of expert-annotated VITILIGO2025 dataset \cite{Guo2022}, which focuses on vitiligo and offers precise segmentation masks validated by dermatologists.

\begin{table}[h]
    \centering
    \caption{Performance on Vitiligo and Retinal Vessel Segmentation Tasks}
    \begin{tabular}{l l c}
        \hline
        Dataset & Model & IoU (\%) \\
        \hline
        \multicolumn{3}{l}{\textit{(a) Vitiligo Segmentation}} \\
        VITILIGO2025 & SegFormer & 76.8 \\
        VITILIGO2025 + ISIC2018 & SegFormer & 80.3 \\
        VITILIGO2025 + ISIC2018 & AFDAN & 90.9 \\
        \hline
        \multicolumn{3}{l}{\textit{(b) Retinal Vessel Segmentation}} \\
        DRIVE & SegFormer & 78.1 \\
        DRIVE + Fundus-AVSeg & SegFormer & 76.5 \\
        DRIVE + Fundus-AVSeg & AFDAN & 82.6 \\
        \hline
    \end{tabular}
    \label{datatril}
\end{table}

\begin{table}[!ht] 
\centering
\scriptsize 
\caption{Performance comparison (in \%) of different segmentation models on VITILIGO2025 and DRIVE datasets.}
\label{iou_out_updated_singlecol}
\resizebox{\linewidth}{!}{
\begin{tabular}{@{}c@{\hspace{5pt}}l@{\hspace{5pt}}cccc@{}}
\toprule
\textbf{Type} & \textbf{Model} & \multicolumn{2}{c}{\textbf{VITILIGO2025}} & \multicolumn{2}{c}{\textbf{DRIVE}} \\
\cmidrule(lr){3-4} \cmidrule(lr){5-6}
& & \textbf{IoU} & \textbf{Dice} & \textbf{IoU} & \textbf{Dice} \\
\midrule
\multirow{5}{*}{\makecell{Traditional\\Model}} 
& PSPNet \cite{PSPNet}        & 78.0 & 87.5 & 68.5 & 81.3 \\
& SAM2 \cite{ravi2024sam2}    & 84.5 & 91.6 & 74.5 & 85.4 \\
& SimCIS \cite{simcis2025}    & 85.8 & 92.4 & 75.2 & 85.8 \\
& CMFormer \cite{CMFormer}    & 87.2 & 93.2 & 76.1 & 86.4 \\
& YOLO11 \cite{ultralytics2024yolo11} & 88.5 & 93.9 & 77.8 & 87.5 \\
\midrule
\multirow{3}{*}{\makecell{Domain\\Adaptation}}
& EHTDI\cite{li2022exploring}         & 89.2 & 94.3 & 79.5 & 88.6 \\
& DSTC-SSDA\cite{gao2024delve}        & 89.8 & 94.6 & 80.2 & 89.0 \\
& \textbf{AFDAN (Ours)}       & \textbf{90.9} & \textbf{95.2} & \textbf{82.6} & \textbf{90.5} \\
\bottomrule
\end{tabular}
}
\end{table}

The second set of experiments evaluates performance on retinal vessel segmentation. We use the Fundus-AVSeg dataset \cite{fundusavseg2025} as the source domain, which provides high-resolution fundus images with pixel-level artery/vein annotations and quality assessment metrics. The target domain is the DRIVE dataset \cite{DRIVE}, a widely recognized benchmark containing fundus images and manually segmented vessel maps, with dual-expert annotations in the test set ensuring fair evaluation of model generalization.

As shown in Table \ref{datatril}, the incorporation of source domain data plays an indispensable role in enabling effective cross-domain knowledge transfer. On both vitiligo and retinal vessel segmentation tasks, our AFDAN model consistently demonstrates the ability to transfer prior knowledge from source to target domains, thereby significantly compensating for the limited annotations in specialized medical image segmentation tasks.

\subsection{Comprehensive Metric-Based Evaluation of Segmentation Models}
We evaluated 8 segmentation models on two medical datasets using core metrics (IoU, Dice; Table \ref{iou_out_updated_singlecol}). Among traditional models, performance showed a consistent upward trend across both datasets: PSPNet \cite{PSPNet} achieved the lowest IoU (78.0\% on VITILIGO2025, 68.5\% on DRIVE), while YOLO11 \cite{ultralytics2024yolo11} emerged as the top traditional model (88.5\% IoU/93.9\% Dice on VITILIGO2025; 77.8\% IoU/87.5\% Dice on DRIVE). Domain adaptation models outperformed traditional ones on both datasets: EHTDI \cite{li2022exploring} and DSTC-SSDA \cite{gao2024delve} reached 89.2\%–89.8\% IoU on VITILIGO2025 and 79.5\%–80.2\% IoU on DRIVE. Our AFDAN achieved the best performance across all metrics and datasets: 90.9\% IoU/95.2\% Dice on VITILIGO2025 (+2.4\% IoU vs. YOLO11) and 82.6\% IoU/90.5\% Dice on DRIVE (+4.8\% IoU vs. YOLO11). These results validate the effectiveness of our proposed method for medical image segmentation.

\subsection{Ablation Studies}

To validate the effectiveness and generalizability of each core component in AFDAN, we conduct ablation studies on both the VITILIGO2025 and DRIVE datasets. Table \ref{tab:ablation} summarizes the IoU (\%) results across both tasks, demonstrating consistent contributions of each module.

\begin{table}[ht]
\centering
\caption{Module effectiveness verification results on vitiligo (VITILIGO2025) and retinal vessel (DRIVE) segmentation.}
\label{tab:ablation}
\begin{tabular}{l c c}
\toprule
Configuration & \multicolumn{2}{c}{IoU (\%)} \\
\cmidrule(lr){2-3}
 & Vitiligo & Retinal Vessel \\
\midrule
Baseline & 80.3 & 76.5 \\
+ STFF & 84.4 (+4.1) & 78.6 (+2.1) \\
+ ADL & 83.9 (+3.6) & 77.4 (+0.9) \\
+ SFI & 85.2 (+4.9) & 78.2 (+1.7) \\
+ STFF + ADL & 86.3 (+6.0) & 79.2 (+2.7) \\
+ STFF + SFI & 88.1 (+7.8) & 81.3 (+4.8) \\
{ \textbf{ Full Model }} & {\textbf{ 90.9} } (+10.6) & {\textbf{ 82.6}} (+6.1) \\
\bottomrule
\end{tabular}
\end{table}

To validate AFDAN’s core components’ effectiveness and generalizability, we perform ablation studies on VITILIGO2025 (vitiligo) and DRIVE (retinal vessel) datasets, assessing each module’s performance contribution. Table \ref{tab:ablation} shows the \textbf{Baseline} (SegFormer without AFDAN’s modules) achieves 80.3\% IoU (vitiligo) and 76.5\% IoU (retinal vessel) as references. Each module boosts performance on both tasks: \textbf{STFF} (frequency alignment for style discrepancy reduction) delivers the largest single-module gains (+4.1\% vitiligo, +2.1\% DRIVE); \textbf{ADL} improves feature invariance to domain shift (+3.6\% vitiligo, +0.9\% DRIVE); \textbf{SFI} (cross-domain attention for structural localization) contributes +4.9\% vitiligo and +1.7\% DRIVE. Notably, module combinations exhibit synergistic effects: \textbf{STFF+SFI} performs exceptionally well (+4.8\% DRIVE over baseline), reflecting complementary global frequency alignment and local attention. The full AFDAN (all modules integrated) achieves the highest IoU: 90.9\% (+10.6\% vitiligo) and 82.6\% (+6.1\% DRIVE), validating the modules’ collaborative enhancement of cross-domain adaptation under annotation scarcity.

\section{Conclusion}
\label{sec:refs}
In this paper, we have proposed AFDAN, a novel cross-domain segmentation network designed to tackle annotation scarcity in medical imaging. The framework integrates three core modules: an Adversarial Domain Learning module that aligns feature distributions via frequency transformation, a Source–Target Frequency Fusion module that accelerates adaptive feature construction, and a Spatial-Frequency Integration module that combines complementary representations to preserve structural details. Extensive experiments show that AFDAN achieves state-of-the-art performance, with a 90.9\% IoU on VITILIGO2025 and 82.6\% IoU on the DRIVE benchmark. The results confirm its robustness in segmenting both normal and pathological regions with high fidelity, demonstrating strong potential for clinical applications where annotated data are limited. 

\bibliographystyle{IEEEbib}
{
\small
\bibliography{refs}
}

\end{document}